\pdfoutput=1

\documentclass[11pt]{article}

\usepackage{ACL2023}

\usepackage{times}
\usepackage{latexsym}

\usepackage[T1]{fontenc}

\usepackage[utf8]{inputenc}

\usepackage{microtype}

\usepackage{inconsolata}
\usepackage{amsfonts}
\usepackage{amsmath}
\usepackage{amssymb}
\usepackage{pifont}
\usepackage{bbold}
\usepackage{graphicx}
\usepackage{subcaption}
\usepackage{color, xcolor}
\usepackage{setspace,placeins}
\usepackage{colortbl}
\usepackage{enumitem}
\usepackage{multirow}
\usepackage{multicol}
\usepackage{makecell}
\usepackage{anyfontsize}
\usepackage{scrextend}

\definecolor{ForestGreen}{RGB}{34,139,34}
\definecolor{FrenchBlue}{rgb}{0.0, 0.45, 0.73}
\definecolor{Brown}{rgb}{0.59, 0.29, 0.0}

\usepackage{kantlipsum}
\usepackage[breakable, theorems, skins]{tcolorbox}
\tcbset{enhanced}

\DeclareRobustCommand{\mybox}[2][gray!20]{%
\begin{tcolorbox}[   
        breakable,
        left=0pt,
        right=0pt,
        top=0pt,
        bottom=0pt,
        colback=#1,
        colframe=#1,
        width=\linewidth, 
        enlarge left by=0mm,
        boxsep=5pt,
        arc=0pt,outer arc=0pt,
        ]
        #2
\end{tcolorbox}
}

\title{\textsc{SyntaxShap}: Syntax-aware Explainability Method for Text Generation}

\author{Kenza Amara,\hspace{.5em}Rita Sevastjanova,\hspace{.5em}Mennatallah El-Assady \\
  Department of Computer Science\\
  ETH Zurich, Switzerland \\
  \normalsize\texttt{\{kenza.amara, menna.elassady\}@ai.ethz.ch}\vspace{-.2em}\\
  \normalsize\texttt{rita.sevastjanova@inf.ethz.ch}}
  
\begin{document}
\maketitle
\begin{abstract}
To harness the power of large language models in safety-critical domains, we need to ensure the explainability of their predictions. However, despite the significant attention to model interpretability, there remains an unexplored domain in explaining sequence-to-sequence tasks using methods tailored for textual data. This paper introduces \textit{SyntaxShap}, a local, model-agnostic explainability method for text generation that takes into consideration the syntax in the text data. The presented work extends Shapley values to account for parsing-based syntactic dependencies. Taking a game theoric approach, SyntaxShap only considers coalitions constraint by the dependency tree. We adopt a model-based evaluation to compare SyntaxShap and its weighted form to state-of-the-art explainability methods adapted to text generation tasks, using diverse metrics including faithfulness, coherency, and semantic alignment of the explanations to the model. We show that our syntax-aware method produces explanations that help build more faithful and coherent explanations for predictions by autoregressive models. Confronted with the misalignment of human and AI model reasoning, this paper also highlights the need for cautious evaluation strategies in explainable AI.\footnote{Code is publicly available \href{https://github.com/k-amara/syntax-shap}{here} and on the \href{https://syntaxshap.ivia.ch/}{SyntaxShap website}.}
\end{abstract}

\section{Introduction}

\begin{figure*}[h]
    \centering
    \includegraphics[width=\linewidth]{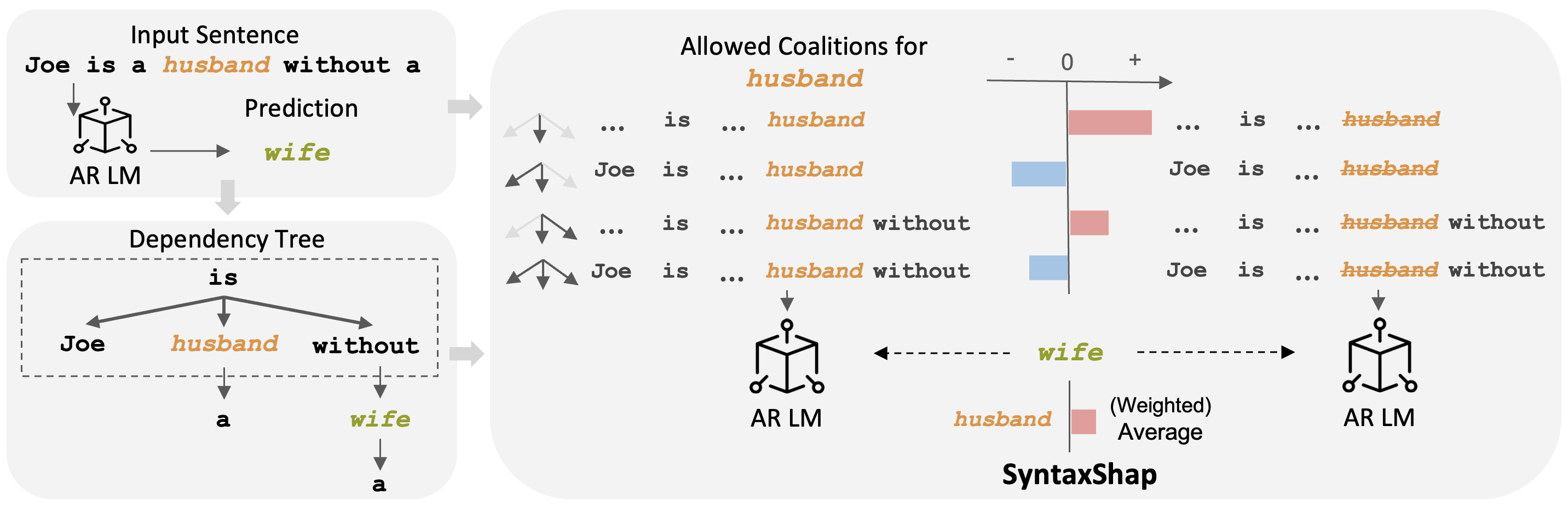}
    \caption{
    Given an input sentence, an autoregressive language model (AR LM) predicts the next token. The syntax of the sentence is extracted using dependency parsing (\texttt{spaCy}~\cite{spacy}). To measure the importance of the word \textcolor{orange}{\textbf{\textit{husband}}} for the model to predict the next token \textcolor{olive}{\textbf{\textit{wife}}}, our method (1) extracts multiple coalitions of words following specific paths in the dependency tree, (2) analyze the contribution of adding \textcolor{orange}{\textbf{\textit{husband}}} to each coalition in the change of probability to predict the next token \textcolor{olive}{\textbf{\textit{wife}}}, and (3) average those contributions to compute its final SyntaxShap value.}
    \vspace{-1em}
    \label{fig:syntaxshap_summary}
\end{figure*}

Language model (LM) interpretability has become very important with the popularity of generative AI. Despite the great results achieved by the most recent LMs, there is still a large range of tasks where the models fail, e.g., capturing negations~\cite{
truong-etal-2023-language}. Therefore, it is crucial to get a better understanding of the LM reasoning and develop faithful explainability methods. As many LMs have little transparency and their use is restricted to API calls, model-agnostic explainability methods have become the most practical techniques for gaining better insights into LMs. 

The SHapley Additive exPlanations (SHAP) framework is popular for generating local explanations thanks to its solid theoretical background and general applicability~\citep{shapley1953value}. However, regarding sequence-to-sequence tasks such as next token generation, the usage of SHAP-based methods has not been explored in depth~\citep{SHAP-NLP}. We address this gap and develop a coalition-based explainability method inspired by Shapley values for text generation explanation. 

Our explainability method (in \autoref{fig:syntaxshap_summary}) considers syntactic word dependencies~\cite{DependencyTree}. The syntax is important as next-word predictions in autoregressive (AR) LMs underlie implicit incremental syntactic inferences, i.e., LMs implicitly capture dependencies in text data~\cite{probingARLM}. 
In this paper, we investigate if dependency parsing trees can be used in the explainability process as syntactic relational graphs and help shed light on the influence of words on the model's prediction given their syntactic role in the sentence.


We evaluate the explanations on diverse metrics. First, we adapt fidelity, one of the most popular model-based evaluation metrics in xAI (eXplainable AI), to the text generation task and introduce two new metrics to test whether the generated explanations are faithful to the underlying model. Second, we introduce two qualitative evaluation metrics that capture the explanation quality with regard to human expectations, i.e., 
the coherency of explanations and their semantic alignment. Our evaluation procedure compares our method \textit{SyntaxShap} to state-of-the-art explainability methods. 
Explanations produced by our method of the next token generation by two popular AR models respect the sentence syntactic structure while being faithful and more coherent compared to state-of-the-art SHAP-based methods that do not explicitly consider the word dependency for text generation tasks. Our semantic alignment analysis also proves that faithful explanations do not necessarily meet human expectations in terms of token importance, highlighting a clear disagreement between our human mental model and AI models.

To summarize, our contributions are (1) SyntaxShap, a new SHAP-based explainability method that incorporates dependency tree information, (2) quantitative metrics that address LM's stochasticity and qualitative metrics to account for human semantic expectations, and (3) an evaluation of the explanation quality on two AR LMs. Our work opens multiple research directions for future work.

\section{Related Work}

\paragraph{Explainability in Linguistics}
Syntax and semantics play an important role in explaining LM outcomes from a linguist perspective. Multiple attempts were made to explore the role of syntactic and semantic representations to enhance LM predictions. \citet{SemanticVSSyntax} look at the role of syntactic and semantic tags for the specific task of human sentence acceptability judgment. They show that syntactic tags significantly influence the predictions of the LM. 
In recent years, there has been an increasing interest in methods that incorporate syntactic knowledge into Machine Translation~\cite{Syntax_machineTranslation}. In addition, \citet{probingARLM} has shown that next-word predictions from AR neural LMs show remarkable sensitivity to syntax. However, there has been no attempt to account for the syntax in explanations of those LMs for text generation tasks~\cite{SHAP-NLP}. For this reason, we propose to incorporate syntax-based rules to explain AR LM text generation.

\paragraph{SHAP-based explainability in NLP} One way to categorize model-agnostic post-hoc explainability methods is to separate them into perturbation-based and surrogate methods~\cite{xAI-NLP}. Among the most popular surrogate models are LIME and SHAP. The Shapley-value approach~\cite{shapley1953value} provides local explanations by attributing changes in predictions for individual data inputs to the model’s features. Those changes can be combined to obtain a better global understanding of the model structure. For text data, available approaches seem mostly tailored to classification settings~\cite{TransShap,HierarchicalTextShapley}. 

\paragraph{Shapley values and complex dependencies} One underlying assumption of SHAP is feature independence. Confronted with more diverse types of data inputs, newer methods offer the possibility to account for more complex dependencies between features. \citet{AsymmetricShapley} propose Asymmetric Shapley values, which drop the symmetry assumption and enable the generation of model-agnostic explanations incorporating any causal dependency known to be present in the data. Following this work, \citet{CausalShapley} propose Causal Shapley values to account more specifically for causal structures behind feature interactions. \citet{LC-Shapley} construct coalitions based on a graph structure, grouping features with their neighbors or connected nodes. When it comes to text data, words present strong interactions, and their 
contribution heavily rely on the context. Therefore, feature attributions for textual data should be specifically tailored to account for those complex dependencies. HEDGE is one example of a SHAP-based method addressing the context dependencies specific to text data~\cite{HierarchicalTextShapley}. It hierarchically builds clusters of words based on their interactions measured by the \textit{cohesion score}. While their objective is to cluster words to minimize the loss of faithfulness, i.e., prediction change, we propose a new strategy to create coalitions of words that respect the syntactic relationships dictated by the dependency tree. This way, we consider the syntactic dependencies that are the basis of linguistics and which were proven essential for next-word predictions from AR LMs~\cite{probingARLM}.

\section{SyntaxShap Methodology}\label{sec:method}

\subsection{Objective}
Given a sentence of $n$ words $\textbf{x} = (x_1,..., x_n)$ and $\hat{\textbf{y}} = (\hat{y}_1,...,\hat{y}_m)$ the $m$ generated words by an AR LM $f$, the objective is to evaluate the importance of each input token for the prediction $\hat{\textbf{y}}$. We focus on explaining the next token, i.e., $m=1$. Let $f_y(x)$ be the model's predicted probability that the input data $\textbf{x}$ has the next token $y$. Our method produces local explanations. 

\subsection{Shapley values approach}

We adopt a game theory approach to measure the importance of each word $x_i$ to the prediction. The Shapley value approach was first introduced in cooperative game theory~\cite{shapley1953value} and computes feature importance by evaluating how each feature $i$ interacts with the other features in a coalition $S$. For each coalition of features, it computes the marginal contribution of feature $i$, i.e., the difference between the importance of all features in $S$, with and without $i$. It aggregates these marginal contributions over all subsets of features to get the final importance of feature $i$.

\subsection{Syntax-aware coalition game}

Our work focuses on incorporating syntax knowledge into model-agnostic explainability. We adopt a coalition game approach that accounts for these syntactic rules. As illustrated in \autoref{fig:syntaxshap_summary}, SyntaxShap computes the contribution of words only considering \textit{allowed} coalitions $\mathfrak{S}$ constraint on the dependency tree structure. We define a coalition $S$ as a set of words or features $\{x_i,i\in [1,n]\}$ from the input sentence $x$. Given a dependency tree with $L$ levels, $l_i \in [1, L]$ corresponds to the level of word $x_i$ in the tree and $n_l>0$ the number of words at level $l$ in the tree. To compute the contribution of the words in the tree, SyntaxShap only considers the allowed coalitions $\mathfrak{S} = \bigcup_{l=0}^{L}\mathfrak{S}_{l}$, where $\mathfrak{S}_{l}$ is the set of allowed coalitions at level $l$. We pose the default $\mathfrak{S}_{0}=\{S_0\}$ and $S_0 = \{\}$ is the null coalition. 

\noindent\textbf{Notations} Let $X_l$ be the set that contains all the words at level $l$, $X_{<l}$ the one that contains all the words before level $l$ in the tree, and $\mathcal{P}(X_l)$ the powerset, i.e. the set of all subsets of $X_l$.

\noindent\textbf{Definition} (Set of coalitions at level $l$) The set of coalitions $\mathfrak{S}_{l}$ at level $l$ is defined as:
\begin{align*}
    \mathfrak{S}_{l} = \bigcup\limits_{\sigma\in\mathcal{P}(X_l)}X_{<l}\cup\sigma
\end{align*}
 
\noindent\textbf{Property} At each level of the tree, each coalition $S\in\mathfrak{S}_l$ respects two properties:
\begin{align}
    \forall i\in [1,n] \text{ s.t. } l_i>l,  x_i\notin S.\\
    \forall i\in [1,n] \text{ s.t. } l_i<l, x_i\in S.
\end{align}

Given the tree-based coalitions, we can compute the contribution of each token in the input sentence to the model's prediction. The contribution of feature $x_i$ at level $l_i$ on the dependency tree to the model output $\hat{y}$ is defined as:
\begin{align}
\phi_{i} = \frac{1}{N_i}\hspace{-1em}\sum_{S\in\left(\bigcup\limits_{p=0}^{l_i-1}\mathfrak{S}_p\right)\bigcup\mathfrak{S}_{l_i}^{\backslash i}}\hspace{-2em}[f_{\hat{y}}(S\cup \{x_i\}) - f_{\hat{y}}(S)]
\end{align}
where $N_i$ corresponds to the number of allowed coalitions at level $l_i$ that do not contain feature $x_i$, and $\mathfrak{S}_{l}^{\backslash i}$ corresponds to the set of coalitions at level $l$ that exclude word $x_i$, i.e., \begin{align*}\mathfrak{S}_{l}^{\backslash i}=\bigcup\limits_{\sigma\in\mathcal{P}(X_l)}X_{<l}\cup(\sigma\backslash \{x_i\}).\end{align*}

\noindent\textbf{Property} Given the number $n_l$ of words (or nodes) at level $l$ of the tree, each word at the same level shares the same number of updates, i.e., allowed coalitions, i.e., $\forall x_i$ s.t. $l_i=l$, $N_i=N^l$ and $N_l$ can be expressed as:
\begin{align}\label{eq:updates}
    N_l = \sum_{p=0}^{l-1} 2^{n_p} + 2^{n_{l}-1} - l
\end{align}
\textbf{Proof} To compute $N_l$ in equation \autoref{eq:updates}, we proceed recursively starting from the root nodes. The dependency has $L$ levels starting from level $l=1$. We postulate a hypothetical level 0 where the null coalition $S_0=\{\}$ can be formed. At level 1, there is the root node of the tree, i.e. $n_1=1$. The number of coalitions is $|\mathfrak{S}_1=\{\{x_{\text{root}}\}\}| = 1$. Let $n_l$ be the number of nodes at level $l$. The number of combinations of $n_l$ features is $2^{n_l}$. Since we already counted the null coalitions at the hypothetical level 0, we don't count it in the allowed coalitions $\mathfrak{S}_l$ at level $l$. We arrive at the final number of coalitions $|\mathfrak{S}_l|=2^{n_l}-1$. 
Now, let's say we have a word $x$ at level $l$. This word can join all allowed coalitions at level $<l$ --- there are $1+\sum_{p=1}^{l-1}(2^{n_p}-1)$ --- and all the coalitions of the words at level $l$ where $x$ does not appear --- there are $2^{n_l-1}-1$. In conclusion, we find that the number of allowed coalitions for word $x$ at level $l$ is:
\begin{align*}
    N_l &= 1 + \sum_{p=1}^{l-1}(2^{n_p}-1) + 2^{n_l-1}-1 \\
    &= \sum_{p=0}^{l-1}2^{n_p} + 2^{n_l-1} - l
\end{align*}
We pose $n_0=0$, the number of nodes on the hypothetical level 0, to start the sum at $p=0$ for simplification.

Our strategy of building tree-based coalitions drops the efficiency assumption of Shapley values but preserves the symmetry axioms for the words at the same level of the dependency tree, as well as the nullity and additivity axioms. Appendix \autoref{apx:axioms} details the four shapley axioms and discusses which ones SyntaxShap respects or violates. Note that this does not undermine the quality of the explanations since the axioms were shown to work against the goals of feature selection in some cases~\cite{AxiomsLimitation}.

\subsection{Weighted SyntaxShap}
In the context of text data and syntactic dependencies, we assume that words at the top of the tree should be given more importance since they are the syntactic foundations of the sentence and usually correspond to the verb, subject, and verb qualifiers. Therefore, we propose SyntaxShap-W, a variant of our method that weighs words according to their position in the tree. The weights are tree-level-dependent and correspond to the inverse of the word level for which contribution is computed, i.e., $w_l = 1/l$. The contribution of a word $x_i$ at level $l_i$ can be expressed as:
\begin{align}
    \phi_{i} = \frac{w_{l_i}}{N_i}\hspace{-1.5em}\sum_{S\in\left(\bigcup\limits_{p=0}^{l_i-1}\mathfrak{S}_p\right)\bigcup\mathfrak{S}_{l_i}^{\backslash i}}\hspace{-2em}[f_{\hat{y}}(S\cup \{x_i\}) - f_{\hat{y}}(S)]
\end{align}

\section{Evaluation}

This section describes our model-based evaluation procedure that encompasses both quantitative and qualitative analysis of the explanations. While previous works only focus on the faithfulness of explanations to assess their quality, we also propose to consider human qualitative expectations.

\subsection{Quantitative evaluation}

To analyze if the explanations are faithful to the model, we adopt \textit{fidelity}, the most common model-based metric in xAI~\cite{fidelity_metric}, which looks at the top-1 prediction and propose two new variants that balance the LM's probabilistic nature by considering the top-K predictions. By considering the first top K predictions, we balance the explainer's deterministic nature and the inherent stochasticity of language models.

\paragraph{Fidelity} Fidelity measures how much the explanation is faithful to the model's initial prediction for the next token. By keeping the top t\% words in the input sentence, fidelity calculates the average change in the prediction probability on the predicted word over all test data as follows,\vspace{-.4em}
\begin{align}
    \text{Fid}(t)=\frac{1}{N}\sum_{i=1}^{N}(f_{\hat{y}}(x_i) - f_{\hat{y}}(\tilde{x}_i^{(t)}))
\end{align}
where $\tilde{x}_i^{(t)}$ is the masked input sentence constructed by keeping the $t$\% top-scored words of $x_i$, $\hat{y}$ is the predicted token given input $x_i$, i.e. $\hat{y} = \mathop{\mathrm{argmax}}\nolimits_{y'}f_{y'}(x_i)$, and N is the number of examples.
Usually, the missing words are replaced by the null token, but we also propose an alternative fidelity Fid$_{rand}$ by replacing the missing words with random words from the tokenizer vocabulary.  

\paragraph{Probability divergence@K} The probability divergence at K corresponds to the average difference in the top K prediction probabilities on the predicted class over all test data. It can be expressed as follows,
\begin{align}
    \text{div@}K = \frac{1}{N}\sum_{i=1}^{N}\sum_{k=0}^K(f_{\hat{y}_k}(x_i) - f_{\hat{y}_k}(\tilde{x_i}^{(t)}))
\end{align}
where $\hat{y}_k$ is the top $k^{\text{th}}$ prediction given input $x_i$. We choose $K=10$ because most of the sentences can be completed with multiple possible words that are synonyms or semantically consistent with the input sentence.

\paragraph{Accuracy@K} The accuracy at K corresponds to the average ratio of common top K predictions between the full and masked sentences:
\begin{align}
    \text{acc@}K = \frac{1}{N}\sum_{i=1}^{N}\frac{\left|\{\hat{y}_k,{\scriptstyle k\leq K}\}\cap\{\tilde{y}_k^{(t)}, {\scriptstyle k\leq K}\}\right|}{K}
\end{align}
where $\tilde{y}_k^{(t)}$ is the top $k^{\text{th}}$ prediction given input $\tilde{x}_i^{(t)}$.

\subsection{Qualitative evaluation}

\paragraph{Coherency} Coherency describes how similar the explanation is w.r.t. similar next generated token. In other words, given a pair of input sentences with a slight variation in the syntax but a strong change in semantics (e.g., differing only by a negation), we expect similar explanations for similar model's predictions and dissimilar ones when the model is sensitive to the perturbation. 

\paragraph{Semantic alignment} An important criterion to evaluate a textual explanation is whether it is aligned with human expectations. As humans, we intuitively expect the language model to draw little attention to tokens in the input sentence, which semantic substance is not reflected in the prediction. This semantic alignment can be measured for some semantically rich tokens that are decisive for text generation, e.g., the negation. Given a decisive token in input sentences and a model's prediction that does not semantically account for it, we compare methods on the importance rank attributed to this token. An explainability method is semantically aligned if this rank is high, i.e., the decisive token is not important for the model's prediction. 

\section{Experiments}

We evaluate SyntaxShap and SyntaxShap-W on various criteria such as faithfulness in section \autoref{sec:faithfulness}, the coherency in section \autoref{sec:coherency}, and the semantic alignment of their explanations in section \autoref{sec:semantic}.

\begin{figure*}
\begin{subfigure}{\textwidth}
    \includegraphics[width=\textwidth,trim={0 2cm 0 0},clip]{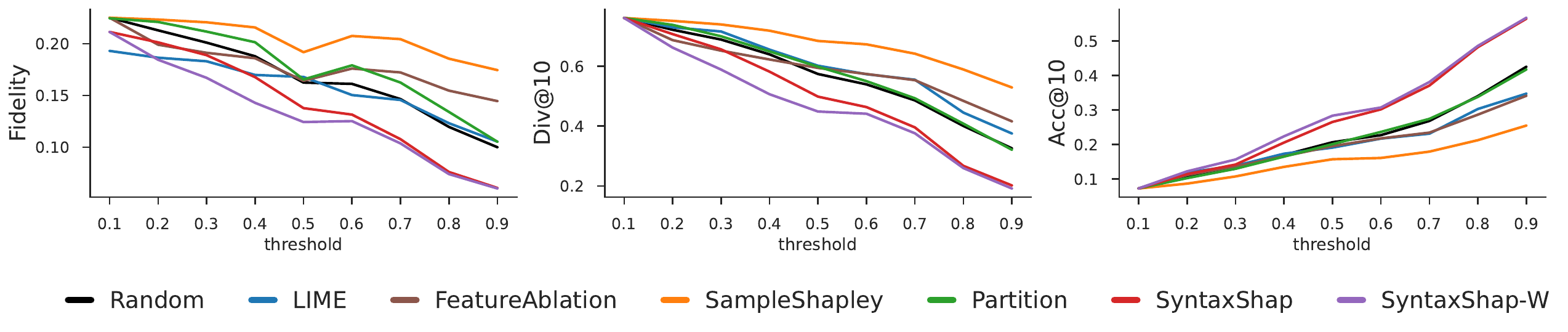}
    \caption{Mistral 7B, Negation dataset.}
    \label{fig:mistral_Negation}
\end{subfigure}
\hfill
\begin{subfigure}{\textwidth}
    \includegraphics[width=\textwidth,trim={0 2cm 0 0},clip]{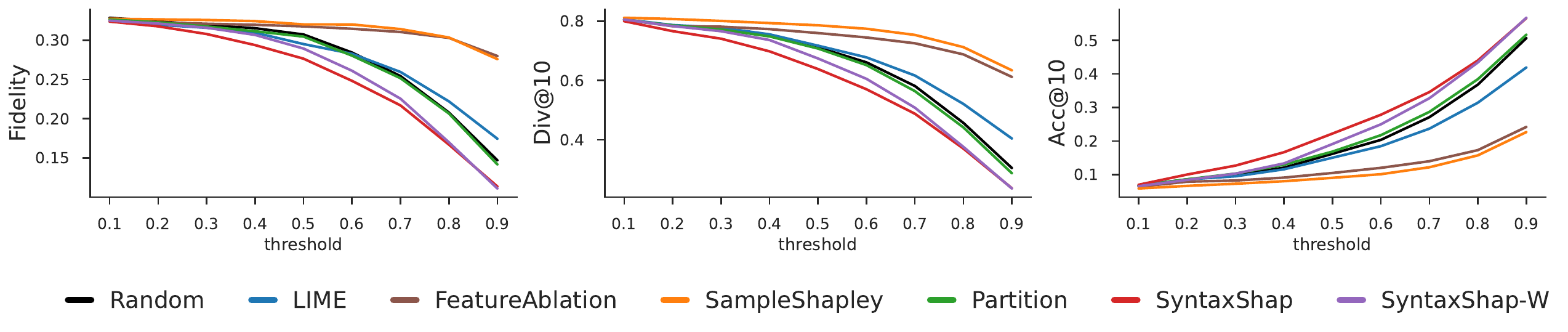}
    \caption{Mistral 7B, Generics dataset.}
    \label{fig:mistral_generics}
\end{subfigure}
\hfill
\begin{subfigure}{\textwidth}
    \includegraphics[width=\textwidth,trim={0 0 0 0},clip]{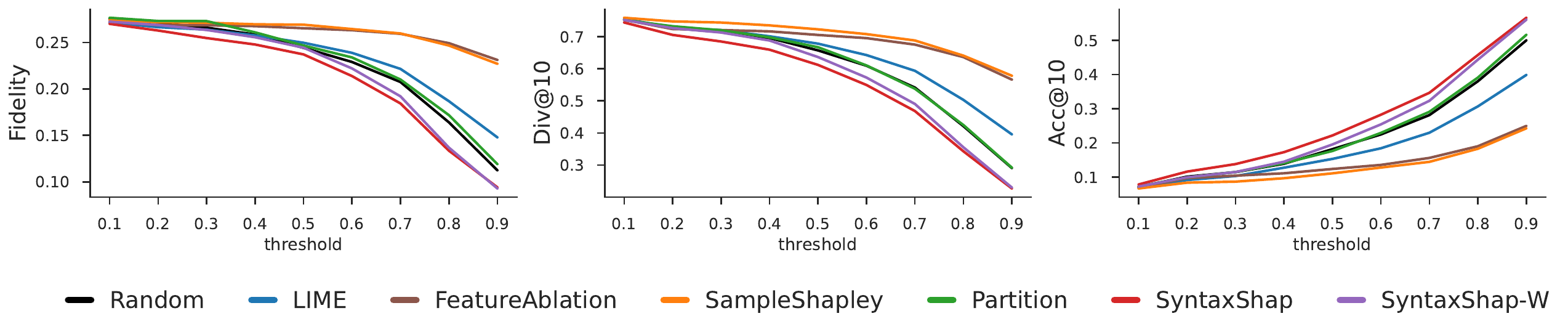}
    \caption{Mistral 7B, ROCStories dataset.}
    \label{fig:mistral_ROCStories}
\end{subfigure}
\caption{Faithfulness of the explanations of Mistral 7B predictions by the methods Random, LIME, FeatureAblation, SampleShapley, Partition, and our methods SyntaxShap and SyntaxShap-W.}
\label{fig:mistral_faithfulness}
\end{figure*}

\begin{figure*}
\centering
\begin{subfigure}{\textwidth}
    \includegraphics[width=\textwidth,trim={0 2cm 0 0},clip]{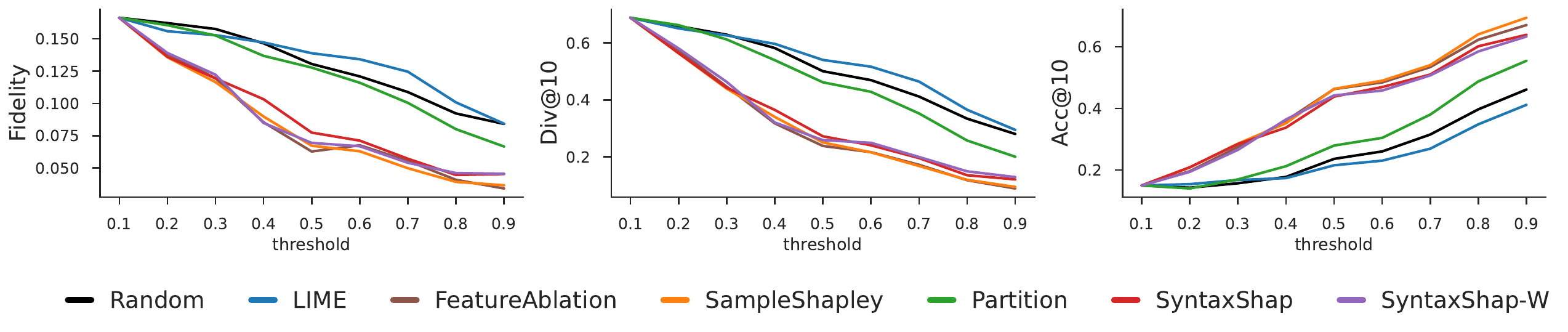}
    \caption{GPT-2, ROCStories dataset}
    \label{fig:gpt2_ROCStories}
\end{subfigure}
\hfill
\begin{subfigure}{\textwidth}
    \includegraphics[width=\textwidth,trim={0 2cm 0 0},clip]{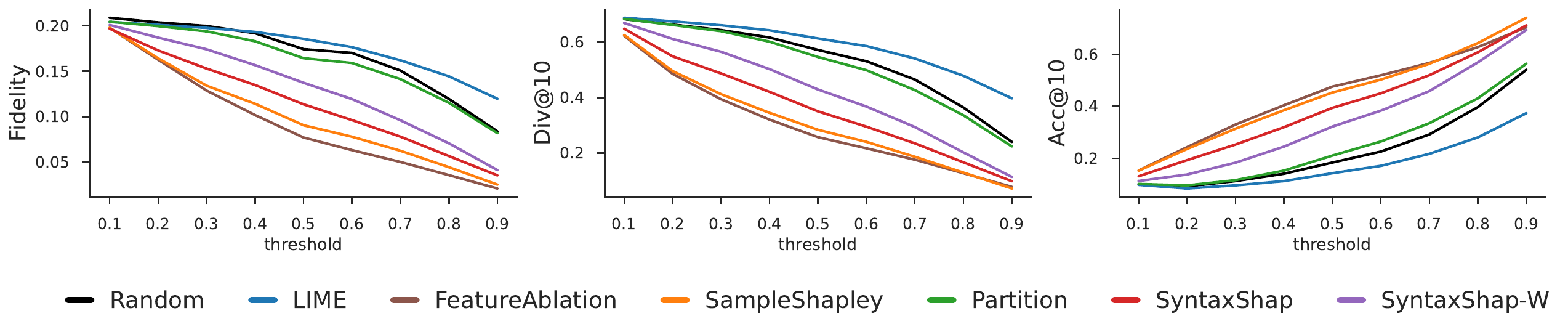}
    \caption{GPT-2, Generics dataset}
    \label{fig:gpt2_generics}
\end{subfigure}
\hfill
\begin{subfigure}{\textwidth}
    \includegraphics[width=\textwidth,trim={0 0 0 0},clip]{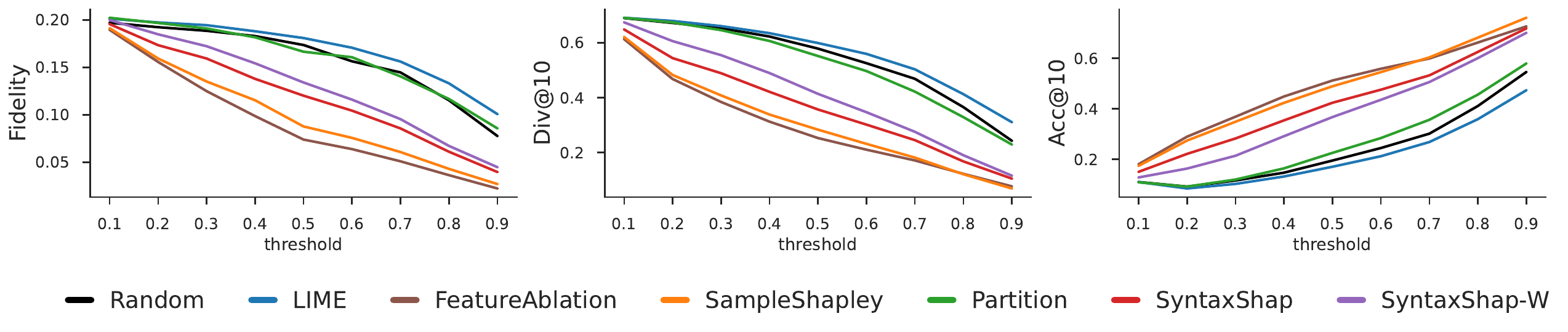}
    \caption{GPT-2, ROCStories dataset}
    \label{fig:gpt2_ROCStories}
\end{subfigure}
\caption{Faithfulness of the explanations of GPT-2 predictions by the methods Random, LIME, FeatureAblation, SampleShapley, Partition, and our methods SyntaxShap and SyntaxShap-W.}
\label{fig:gpt2_faithfulness}
\end{figure*}

\begin{table*}
\centering
    \small
        \begin{tabular}{lccc|ccc}
        & \multicolumn{3}{c}{\textbf{\normalsize{Mistral 7B}}} & \multicolumn{3}{c}{\textbf{\normalsize{GPT-2}}} \\
        & \textbf{Negation} & \textbf{Generics} & \textbf{ROCStories} & \textbf{Negation} & \textbf{Generics} & \textbf{ROCStories}\\\hline
        Random & $0.574$\scriptsize$\pm 0.002$ & $0.711$\scriptsize$\pm 0.006$ & $0.657$\scriptsize$\pm 0.002$ & $0.501$\scriptsize$\pm 0.003$ & $0.572$\scriptsize$\pm 0.003$ & $0.578$\scriptsize$\pm 0.002$\\
        LIME & $0.577$\scriptsize$\pm 0.006$ & $0.689$\scriptsize$\pm 0.003$ & $0.660$\scriptsize$\pm 0.004$ & $0.555$\scriptsize$\pm 0.007$ & $0.590$\scriptsize$\pm 0.003$ & $0.591$\scriptsize$\pm 0.002$\\
        FeatureAblation & $0.594$\scriptsize$\pm 0.003$ & $0.760$\scriptsize$\pm 0.003$ & $0.705$\scriptsize$\pm 0.007$ & $\textbf{0.239}$\scriptsize$\pm 0.001$ & $\textbf{0.257}$\scriptsize$\pm 0.005$ & $\textbf{0.253}$\scriptsize$\pm 0.007$\\
        SampleShapley & $0.684$\scriptsize$\pm 0.005$ & $0.786$\scriptsize$\pm 0.002$ & $0.723$\scriptsize$\pm 0.004$ & $0.251$\scriptsize$\pm 0.004$ & $0.283$\scriptsize$\pm 0.005$ & $0.284$\scriptsize$\pm 0.007$\\
        Partition & $0.599$\scriptsize$\pm 0.001$ & $0.708$\scriptsize$\pm 0.005$ & $0.667$\scriptsize$\pm 0.007$ & $0.462$\scriptsize$\pm 0.001$ & $0.546$\scriptsize$\pm 0.004$ & $0.551$\scriptsize$\pm 0.006$\\
        \rowcolor{blue!10}
        SyntaxShap & $0.499$\scriptsize$\pm 0.002$ & $\textbf{0.638}$\scriptsize$\pm 0.007$ & $\textbf{0.612}$\scriptsize$\pm 0.005$ & $0.273$\scriptsize$\pm 0.002$ & $0.350$\scriptsize$\pm 0.002$ & $0.357$\scriptsize$\pm 0.002$\\
        \rowcolor{blue!10}
        SyntaxShap-W & $\textbf{0.449}$\scriptsize$\pm 0.001$ & $0.674$\scriptsize$\pm 0.002$ & $0.637$\scriptsize$\pm 0.004$ & $0.259$\scriptsize$\pm 0.005$ & $0.429$\scriptsize$\pm 0.002$ & $0.414$\scriptsize$\pm 0.005$\\\Xhline{3\arrayrulewidth}
        \end{tabular}
        \caption{The div@10 scores of explainability methods for the GPT-2 and the Mistral 7B model. Explanations are sparse at threshold $t=0.5$, i.e. we keep 50\% of the top words. We report the mean and variance after running experiments on four different random seeds. The methods introduced in this paper are\,\colorbox{blue!10}{shaded.}}

\label{tab:div_at_10_mistral}
\end{table*}

\subsection{Experimental setting}

For the evaluation, we use three datasets, i.e., the \textit{Generics KB}\footnote{\label{ccby4}published at the ACL Anthology, CC BY 4.0 License} (\textit{Generics}) \cite{huggingface:dataset}, \textit{ROCStories Winter2017}\footnote{publicly available, no license} (\textit{ROCStories}) \cite{mostafazadeh-etal-2017-lsdsem}, and \textit{Inconsistent Dataset Negation}\footref{ccby4} (\textit{Negation}) \cite{kalouli-etal-2022-negation}. They have the following characteristics: (1) The \textit{Generics} dataset contains high-quality, semantically complete statements; (2) The \textit{ROCStories} dataset contains a collection of five-sentence everyday life stories; (3) The \textit{Negation} dataset contains disjoint sentence pairs, i.e., a sentence and its negated version. 
For evaluation purposes, we first separate the stories of the \textit{ROCStories} dataset into single sentences and remove the last token from sentences in the three datasets.
We use the TextDescriptives component in \texttt{spaCy} to measure the dependency distance of the analyzed sentences following the universal dependency relations established by~\citet{DependencyTree} and compute the average number of tokens per sentence as well as the number of unique tokens in the three datasets. As shown in \autoref{tab:dataset_descriptives}, sentences in the \textit{Generics} and \textit{ROCStories} datasets have more complex syntactic structures, and the sentences are longer than in the \textit{Negation} dataset. However, the Negation dataset includes sentences with minimal syntactic variation but great semantic differences, enabling fine-grained qualitative analysis. This makes it the most suitable dataset for comparing xAI methods on coherency and semantic alignment.

\begin{table}
    \centering
    \small
    \begin{tabular}{lccc}
        \Xhline{3\arrayrulewidth}
         & \textbf{Generics} & \textbf{ROCStories} & \textbf{Negation} \\\hline
        Depd. Dist. $\mu$ & 1.96 & 2.12 & 1.4  \\
        Depd. Dist. $\sigma$ & 0.46 & 0.47 & 0.30  \\
        \# Tokens Mean & 9.80 & 9.83  & 5.54 \\
        \# Unique Tokens & 3548  & 2082 & 99 \\\Xhline{3\arrayrulewidth}
    \end{tabular}
    \caption{Dataset descriptives.}
    \vspace{-1.5em}
    \label{tab:dataset_descriptives}
\end{table}
To assess the performance of our method, we use two AR LMs: GPT-2 model \cite{radford2019language} consisting of 154M parameters~\footnote{GPT-2 model was taken from HuggingFace \url{https://huggingface.co/openai-community/gpt2}} and Mistral 7B \cite{jiang2023mistral} with 7B parameters~\footnote{Mistral 7B model was taken from HuggingFace \url{https://huggingface.co/mistralai/Mistral-7B-v0.1}}. This significant increase in parameters allows Mistral 7B to generate next tokens that are more contextualized and semantically rich. Mistral 7B was also shown superior to other 7B LMs and therefore much more advanced than GPT-2 \cite{jiang2023mistral}. For this reason, we favor Mistral 7B for the qualitative analysis, as the predictions are more specific and better aligned with the context of the input sentences. We reproduce our experiments on four different seeds and convey the mean and variance of our results.
Our methods SyntaxShap and SyntaxShap-W are compared against the \textit{Random} baseline, i.e. a normally distributed token importance attribution, and two other explainability baselines \textit{LIME}~\cite{LIME} and \textit{FeatureAblation}, a perturbation-based method that replaces each input feature with a given baseline and computes the difference in output. FeatureAblation's individual token substitutions may not fully reflect feature interactions. We also compare to \textit{SampleShapley}, an approximation of SHAP that computes the contribution of each input token considering random permutations of the input features. All baselines have been adapted for text data and LLMs by defining interpretable text features and handling sequential predictions. We also compare them against \textit{Partition}, a faster version of KernelSHAP that hierarchically clusters features. Its implementation is based on HEDGE~\cite{HierarchicalTextShapley}, a SHAP-based method that builds hierarchical explanations via divisive generation, respecting some pre-computed word clustering, and is particularly suited for text data\footnote{The code for FeatureAblation and SampleShapley was taken from the \href{https://captum.ai/api/index.html}{Captum} Python library, and for Partition from the \href{https://github.com/shap/shap/blob/master/docs/index.rst}{SHAP} Python library. LIME was adapted for text data from its \href{https://github.com/marcotcr/lime}{initial implementation.}}.

To derive textual explanations from the explanatory masks, we opted for consistency with the SHAP Python library by adopting attention mask modification. This method involves setting the attention weight of a token to zero to simulate its removal from the input data. In our evaluation with faithfulness metrics, we experimented with various strategies, including modifying attention weights and replacing tokens with random selections from the tokenizer vocabulary. However, we found minimal differences in the results across these approaches. For a detailed comparison of these masking strategies, we direct readers to the Appendix \autoref{apx:masking}.

\subsection{Faithfulness}\label{sec:faithfulness}

In this section, we evaluate the faithfulness of our explanations to Random, LIME, FeatureAblation, SampleShapley, Partition, and our methods SyntaxShap and SyntaxShap-W on the full datasets with sentence lengths between 5 and 15 tokens.

For both models, Mistral 7B in \autoref{fig:mistral_faithfulness} and GPT-2 in \autoref{fig:gpt2_faithfulness}, our methods SyntaxShap and SyntaxShap-W produce more faithful explanations than the trivial random algorithm, the LIME method adapted to NLP tasks, and Partition, the state-of-the-art shapley-based local explainability method for text data. Therefore, building coalitions based on syntactic rules gives more faithful explanations than when minimizing a cohesion score, preserving the strongest word interactions~\cite{HierarchicalTextShapley}. For the Mistral 7B model, they also both outperform the leave-one-out method, FeatureAblation, and the approximation-based shapley values, SampleShapley, which produce poorly faithful explanations of Mistral 7B predictions. For GPT-2 predictions on the Generics and ROCStories datasets in \autoref{fig:gpt2_generics} and \autoref{fig:gpt2_ROCStories}, SyntaxShap(-W) still lags slightly behind FeatureAblation and SampleShapley. However, while generating more faithful explanations for GPT-2 model, FeatureAblation and SampleShapley's ablate tokens independently, overlooking highly correlated tokens within a sentence~\cite{CaptumLM}.

\mybox{SyntaxShap(-W) generates more faithful explanations than the random baseline, LIME, and Partition. Although it does not beat the baselines FeatureAblation and SampleShapley for every language model, it considers the intrinsic nature of text data, accounting for token correlations and syntactic structure.}

\subsection{Coherency}\label{sec:coherency}

\begin{figure}[h]
    \centering
    \includegraphics[width=
    \linewidth]
    {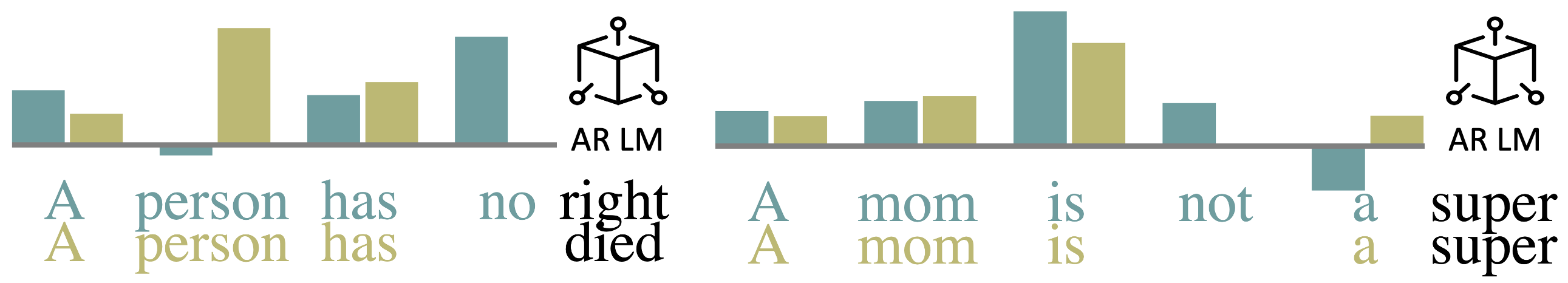}
    \caption{An example of attribution values of the SyntaxShap-W method for two sentence pairs with different and similar next token predictions.}
    \label{fig:sentence-pair}
\end{figure}

In this section, we explore whether SyntaxShap produces coherent explanations with the model understanding. 
For this evaluation, we use Mistral 7B and run a perturbation analysis using sentence pairs from the \textit{Negation} dataset.
We use a sample of 267 sentence pairs (with and without the negation \textit{not} and with varying usage of \textit{with} and \textit{without}) whereby for 90 pairs, the model predicts the same next token. An example of two sentence pairs is shown in \autoref{fig:sentence-pair}. For pairs with equal predictions (e.g., \textit{A mom is \textbf{not} a} and \textit{A mom is a} with an equal next token prediction \textbf{super}), we expect more similar attribution ranks than for pairs with different predictions (e.g., \textit{A person has \textbf{no}} \textbf{right} and \textit{A person has} \textbf{died}). 
\begin{figure}[h!]
    \centering
    \includegraphics[width=\linewidth]{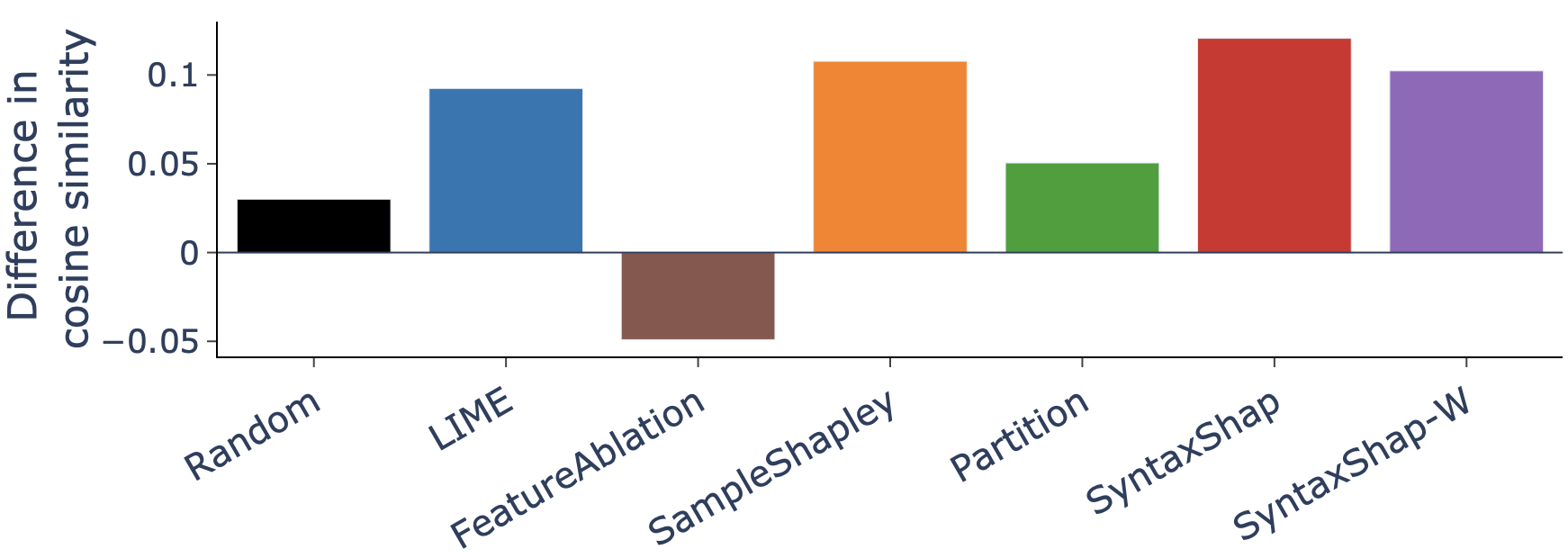}
    \caption{Coherency of explainability methods for the Mistral 7B model on sentence pairs varying by the used negation. SyntaxShap and SyntaxShap-W produce more similar attribution scores for sentence pairs where the model predicts the same next token compared to sentence pairs with different next token predictions.}
    \label{fig:coherency}\vspace{-0.5em}
\end{figure}
To evaluate the coherency, we first represent the attribution scores as rank vectors. 
We then separate pairs with equal predictions and different predictions into two distinct groups and measure the cosine similarity between rank vectors of each pair within each group, whereby negation words are excluded to get equal-length vectors. 
The average difference in cosine similarity between the two groups for each explainability method is displayed in \autoref{fig:coherency}.
It shows that SyntaxShap produces more similar attributions for sentence pairs that predict the same next token and more diverse attributions for sentence pairs with different next token predictions. 
\mybox{Given a pair of sentences with and without a negation, which theoretically have two disjoint semantic meanings, the similarity of SyntaxShap's token attribution values for each sentence better reflect the degree of similarity of the next token predictions than LIME, FeatureAblation, SampleShapley and Partition.}

\subsection{Semantic alignment}\label{sec:semantic} 

\begin{figure}[!h]
    \centering
        \includegraphics[width=\linewidth, trim=0 0 0 22, clip]{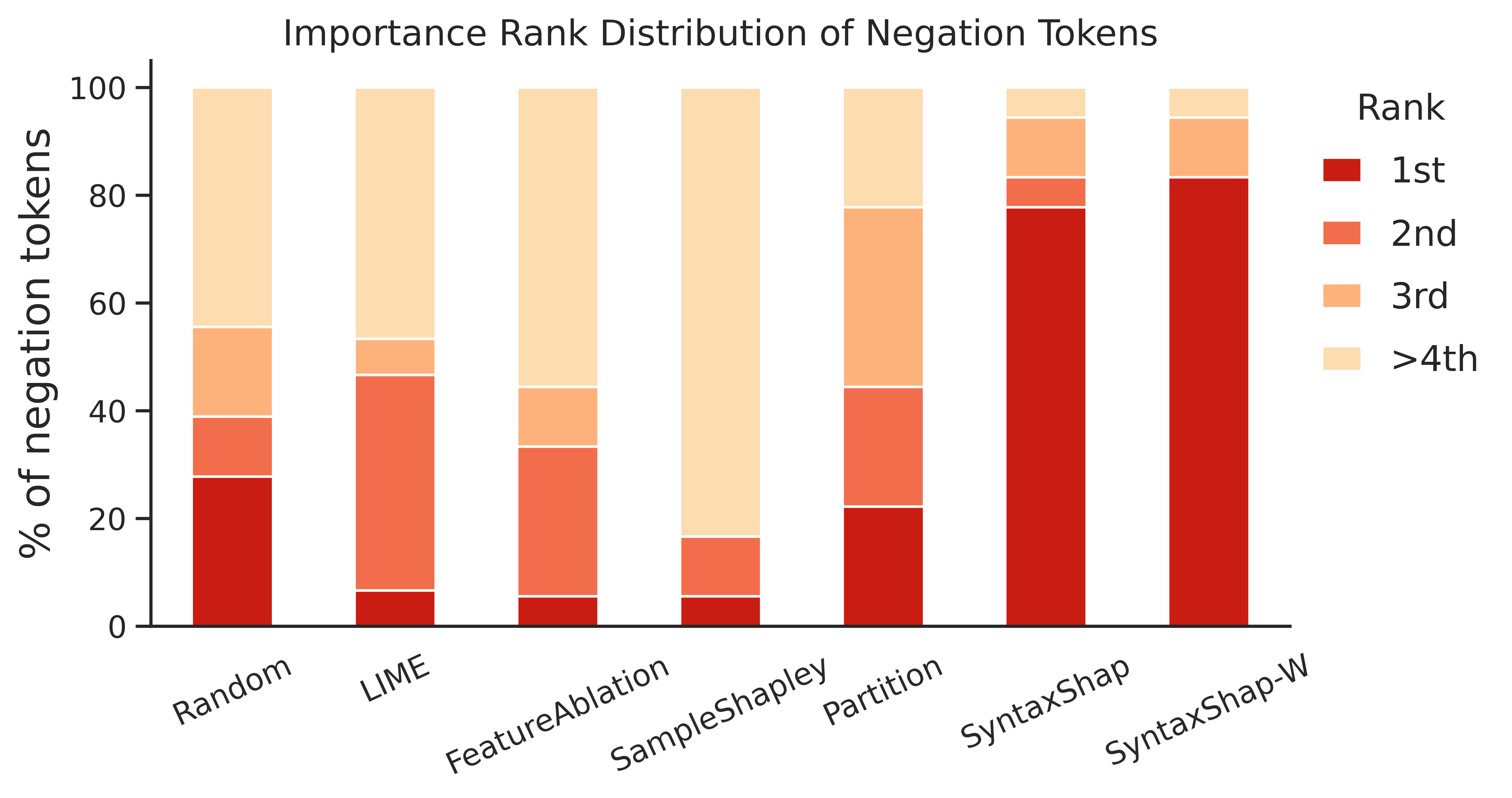}
    \caption{Importance rank distribution of negation tokens \textit{not}, \textit{no}, and \textit{without} when Mistral 7B model does not capture the negations in the predicted next token.\label{fig:semantic_align_mistral}}
\end{figure}

Here, we explore whether the generated explanations are aligned with human semantic expectations. We analyze cases where there is a negation in a sentence, but the model's prediction does not reflect it, e.g., \textit{A father is not a \textbf{father}}. We extract negative instances, i.e., that contain the token \textit{not}, \textit{no}, or \textit{without}, from the \textit{Negation} dataset. We label those where the language model predicts \textit{wrong} next tokens, i.e., semantically misaligned with the negation. The average importance scores of the negation tokens in each of the 22 labeled instances for Mistral 7B are reported in \autoref{fig:semantic_align_mistral}. Following our human mental model, we expect a low importance score assigned to the negation token if the model produces predictions that do not account for it.
However, \autoref{fig:semantic_align_mistral} shows that SyntaxShap(-W) identifies negation as the most important token in more than 80\% of the cases. This indicates a misalignment between Mistral 7B's reasoning and our mental model. Furthermore, SyntaxShap(-W) produces the most faithful explanations for the Negation dataset with Mistral 7B (see \autoref{fig:mistral_faithfulness}). This suggests that faithful explanations do not match human expectations. The misalignment between human and AI reasoning supports the belief that model-focused explanations, which are faithful to the model, are not necessarily intended for human interpretation. Therefore, the evaluation of explanations depends on the focus: human-focused explanations must meet human expectations, while model-focused explanations are intended to be faithful to the language model.  

\mybox{Explainability methods, although faithful, can assign low-importance scores to tokens that humans consider decisive for predictions, revealing a fundamental misalignment between human and AI model reasoning.}

\section{Discussion}
\vspace{-.5em}

\paragraph{Addressing stochasticity} The traditional faithfulness metrics like fidelity, AOPC~\cite{AOPC_2017,AOPC_2018} or log-odds~\cite{Log-odds2017,LC-Shapley} scores take a deterministic approach to evaluate explanations computed on stochastic predictions. 
This paper evaluated AR LMs that adopt top-k sampling to randomly select a token among the k tokens with the highest probability. To account for this stochasticity, we proposed additional evaluation metrics, div@K and acc@K, that consider not only the final prediction but the top-K predictions, balancing the model's probabilistic nature. Nevertheless, further methods that address the stochastic nature of the models should be designed in future research.

\paragraph{Integrating linguistic knowledge} To ensure that the explainability methods produce meaningful explanations that mimic AR LM behavior, we need to go beyond the faithfulness type of evaluation and consider further explainability aspects. In this paper, we study explanations on other dimensions related to semantic interpretation and coherency of explanations. There is potential for more linguistically tailored evaluation methods in the future. The motivation is as follows. The next token prediction task can be seen as a multi-class classification with a large number of classes. The classes have diverse linguistic properties, i.e., tokens have different roles in the sentence, some being more content- and others function-related. We might want to consider these different roles when evaluating the quality of explanations. On the one hand, with controlled perturbations on the input sentences, we can evaluate the role of semantics and syntax on the next token prediction task. 
On the other hand, when computing the explanation fidelity, we might consider prediction changes from one category of tokens (e.g., function words) to another (e.g., content words), giving a more linguistic-aware explanation quality assessment. 

\paragraph{Considering humans} When designing evaluation methods, we need to consider humans since, ideally, they should understand model behavior from the produced explanations. There is one main concern, though. As prior work has shown~\cite{SeEl2022Beware}, LM explainability can suffer from human false rationalization of model behaviors. 
We typically expect the explanations to align with our mental models of language. However, LMs learn language differently from humans; thus, explanations can theoretically differ from our expectations. Thus, future work should design evaluation methods that clearly show the importance of the words for the model and the reasons why this importance (potentially) does not align with human expectations.

\section{Conclusion}\vspace{-.5em}
We proposed SyntaxShap - a local, model-agnostic syntax-aware explanability method. Our method is specifically tailored for text data and meant to explain text generation tasks by AR LMs, whose interpretability in that context has not yet been addressed. SyntaxShap is the first SHAP-based method to incorporate the syntax of input sentences by constraining the construction of word coalitions on the dependency trees. Our experimental results demonstrate that SyntaxShap and its weighted variant can improve explanations in many aspects: they generate more faithful and coherent explanations than the standard model-agnostic explainability methods in NLP for advanced AR LMs like Mistral 7B. The semantic alignment analysis also shows that faithfulness and human intelligibility are distinct evaluation criteria that do not necessarily align. This study addresses a pressing and significant issue regarding the explainability of AR models, contributing to an ongoing dialogue in the research community. 

\newpage
\section*{Limitations}\label{sec:limitations}

\paragraph{Multiple sentences} In this paper, our analysis is limited to one input sentence because we work on one dependency tree at a time. However, our method can be scaled to text with multiple sentences or a paragraph by breaking it down into multiple dependency trees and running SyntaxShap in parallel. However, by doing this, we might lose sentence correlations.

\paragraph{Incorrect dependency tree} Our method heavily relies on the dependency tree, assuming it correctly captures the syntactic relationships between the words. However, the Python module \texttt{spaCy} sometimes generates arguable dependencies from the perspective of linguists, and its accuracy drops when implemented for languages other than English. Therefore, SyntaxShap is, for now, only meant to be used for English grammatically non-convoluted sentences to limit the uncertainty coming from the construction of the dependency tree.


\section*{Ethics Statement}

The data and resources utilized in this study are openly available and widely used by numerous existing works. The datasets employed consist of factual statements devoid of subjective judgments or opinions. It is acknowledged that pre-trained LMs, such as GPT-2 and Mistral 7B, may inherently inherit biases as highlighted in previous research~\cite{radford2019language}, potentially influencing the generated next token. For example, certain tokens like \textit{beautiful} may tend to appear more frequently in contexts associated with female characteristics. While the primary objective of this study is to produce explanations that faithfully represent the model's predictions, it is recognized that these explanations may also carry inherent biases. It is imperative to acknowledge that the results generated by our approach may not always align with human mental models and could potentially be used in applications that have the potential to cause harm.



\bibliography{bibliography}
\bibliographystyle{acl_natbib}
\newpage
\appendix
\section{Textual data}

\subsection{Text generation}
 Text generation tasks involve predicting the next word in a sequence, like in language modeling, which can be considered a simpler form of text generation. Other tasks may involve generating entire paragraphs or documents. Text generation can also be framed as a sequence-to-sequence (seq2seq) task that aims to take an input sequence and generate an output sequence for machine translation and question answering. AR models like GPT (Generative Pre-trained Transformer) generate text one word at a time in an AR manner, conditioning each word on the previously generated words. In this paper, we focused on the next token generation task given one single sentence as input. We work with factual sentences from \textit{Generics} and \textit{ROCStories} datasets, which often expect a semantically rich final token to complete the clause. Multiple predictions are possible, but only a few are correct. Here is an example of a sentence in the \textit{Generics} dataset: \textit{Studio executive is an employee of a film}. The GPT-2 model predicts \textit{studio} as the next token with the random seed 0. We can expect other predictions like \textit{company}, \textit{firm}, or \textit{corporation}. But the number of possibilities is still very limited.

\subsection{Dependency parsing}
Dependency parsing is a natural language processing technique that involves analyzing the grammatical structure of a sentence to identify the relationships between words~\cite{DependencyTree}. It involves constructing a tree-like structure of dependencies, where each word is represented as a node, and the relationships between words are represented as edges. Each relationship has one head and a dependent that modifies the head, and it is labeled according to the nature of the dependency between the head and the dependent. These labels can be found at Universal Dependency Relations~\cite{DependencyTree}. Dependency Parsing is a powerful technique for understanding the meaning and structure of language and is used in various applications, including text classification, sentiment analysis, and machine translation. We use the Python module \texttt{spaCy} (version 3.7.2) \cite{spacy} to generate dependency trees on the input sentences. The number of tokens varies from 5 to 15 tokens for the \textit{Generics} and \textit{ROCStories} datasets, producing very diverse and complex parsing trees. This diversity enriches our analysis and strengthens our results. The code source for dependency parsing was extracted from \url{https://stackoverflow.com/questions/7443330/how-do-i-do-dependency-parsing-in-nltk}.

\section{SyntaxShap: characteristics and proofs}\label{apx:proofs}

\subsection{SyntaxShap and the Shapley axioms}\label{apx:axioms}

The four axioms satisfied by Shapley values, i.e., efficiency, additivity, nullity, and symmetry, do not generally provide any guarantee that the computed contribution value is suited to feature selection, and may, in some cases, imply the opposite~\cite{AxiomsLimitation}. We define here new axioms for SyntaxShap values since two of the four Shapley axioms cannot be satisfied by tree-constraint values.
\paragraph{\textit{Efficiency}} The evaluation function $v(S)$ in SyntaxShap is the output probability for the predicted next token given the full input sentence, i.e., $v(S) = f_{\hat{y}}(S)$ where $\hat{y}=$argmax$(f(x))$. Because of the non-linearity of LMs, SyntaxShap evaluation function is non-monotonic. It does not necessarily increase if you add more features. For this reason, SyntaxShap does \textit{not} satisfy the \textit{efficiency} axiom. This implies that the SyntaxShap values of each word do not sum up to the SyntaxShap value of the whole sentence.
\paragraph{\textit{Symmetry}} SyntaxShap satisfies the axiom of \textit{symmetry} at each level of the dependency tree. Any two features $x_i,x_j$ that are at the same level of the dependency tree, i.e., $l_i=l_j$, play equal roles and therefore have equal SyntaxShap values: 
\begin{align}
    \forall i,j \text{ s.t. }l_i=l_j\notag\\
    [\forall (S\setminus \{x_i,x_j\}) v(S\cup{x_i}) = v(S\cup{x_i})] \notag\\
    \implies \phi_i=\phi_j
\end{align}
\paragraph{\textit{Nullity}} If feature $x_i$ contributes nothing to each submodel it enters, then its SyntaxShap value is zero.
\begin{align}
[(\forall S) v(S\cup\{x_i\}) = v(S)] \implies \phi_i = 0
\end{align}

\paragraph{\textit{Additivity}} Given two models $f$ and $g$, the SyntaxShap value of those models is a linear combination of the individual models’ SyntaxShap values:
\begin{align}
\phi_i(f+g) = \phi_i(f)+\phi_i(g)
\end{align}

\begin{figure*}[t]
\centering
\begin{subfigure}{.48\textwidth}
    \includegraphics[width=0.85\linewidth,trim={0 0 5cm 0},clip]{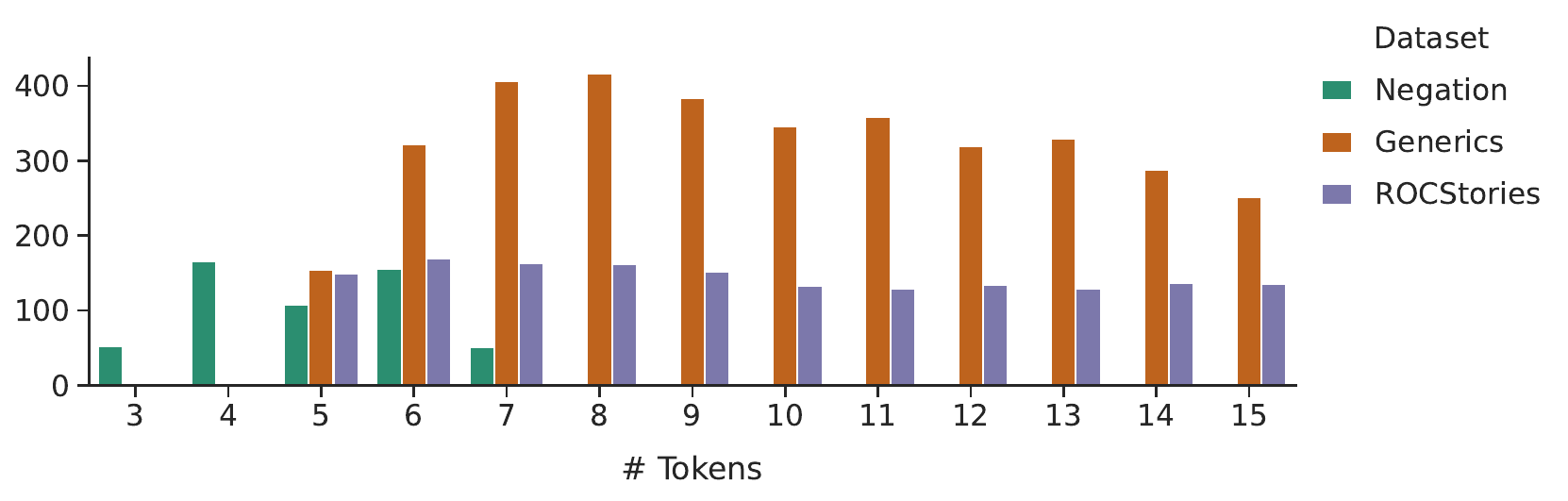}
    \caption{Filtering with GPT-2 model.}
    \label{fig:data_stats_gpt2}
\end{subfigure}
\begin{subfigure}{.48\textwidth}
    \includegraphics[width=\linewidth]{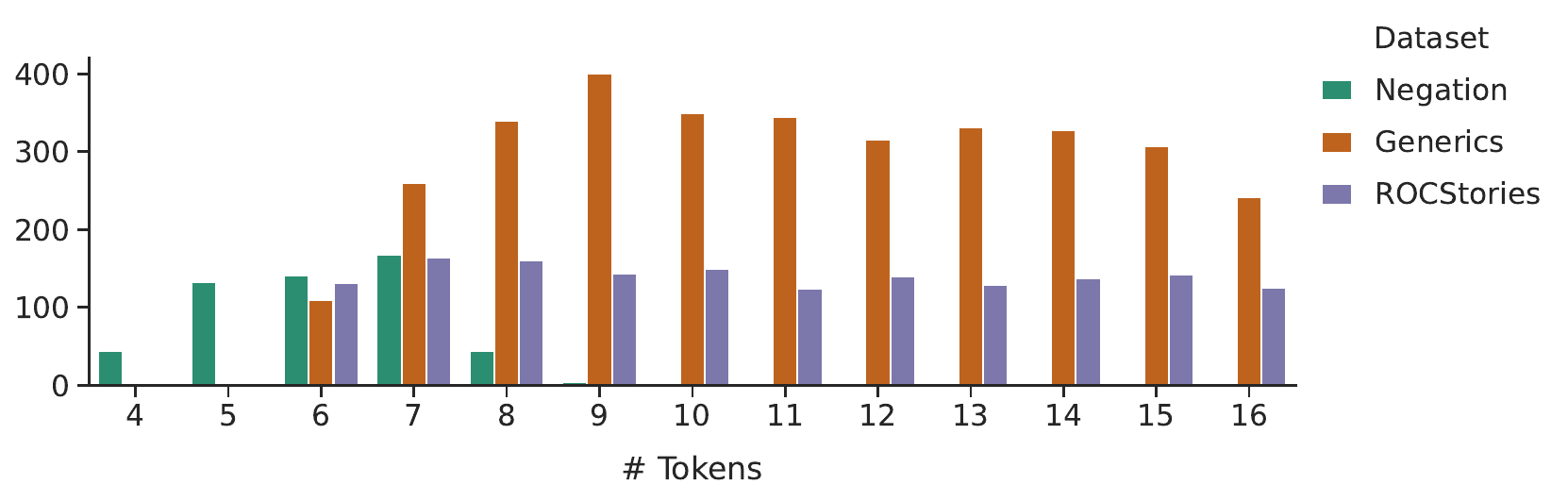}
    \caption{Filtering with Mistral 7B model.}
    \label{fig:data_stats_mistral}
\end{subfigure}
\caption{Number of tokens distribution for the three datasets \textit{Inconsistent Negation Dataset} (Negation), \textit{Generics KB} (Generics) and \textit{ROCStories Winter2017} (ROCStories).}
\label{fig:n_token_count}
\end{figure*}

\subsection{Computational complexity}\label{apx:complexity}

One advantage of the SyntaxShap algorithm is its faster computation time compared to the naive Shapley values computations. We estimate each algorithm's complexity by approximating the total number of computation steps, i.e., formed coalitions and updated values, for the original Shapley values computation and our method.
\paragraph{Shapley values computation} The Shapley value of feature $x$ requires the $2^{n-1}$ coalitions of all features excluding $x$. As we need to update $n$ features, the total number of updates is $n\cdot2^{n-1}$. The computation complexity is, therefore, in $\mathcal{O}(n2^n)$.
\paragraph{SyntaxShap} The SyntaxShap value of feature $x$ at level $l$ requires $N_l$ updates. Considering all the features in the input, the total number of computations is $\sum\limits_{l=1}^L n_l\cdot N_l$. To approximate this number, we assume the dependency tree is balanced and poses $n_l=n/L$. In this case, $N_l$ can be re-written as:
\begin{align*}
    N_l &= \sum_{p=0}^{l-1}2^{n/L} + 2^{n/L-1} - l\\
    &= l(2^{n/L}-1)+2^{n/L-1}
\end{align*}
The total number of computations can now be approximated to:
\begin{align*}
    \frac{n}{L}\sum_{l=1}^L N_l &= \frac{n}{L}\sum_{l=1}^L\left(l(2^{n/L}-1)+2^{n/L-1}\right)\\
    &=\frac{n}{L}\left(\frac{L(L+1)}{2}(2^{n/L}-1) + L2^{n/L-1}\right)\\
    &=\frac{n(L+1)}{2}(2^{n/L}-1) + n2^{n/L-1}
\end{align*}
The approximation of the computation complexity in the case of a balanced tree is $\mathcal{O}(nL2^{n/L})$.

\section{Data preprocessing}\label{apx:filtering}

SyntaxShap relies on the construction of dependency trees that capture the syntactic dependencies in the sentences. Entities in dependency trees are words as defined in the English dictionary. However, language tokenizers sometimes split words into multiple tokens. SyntaxShap supports subtokenization, which involves splitting a word into multiple tokens. When this occurs, SyntaxShap duplicates the word-node in the dependency tree, ensuring that each token is viewed as a distinct node with an identical role. Subtokens receive equal treatment in the analysis process. \autoref{tab:dataset_size} displays the statistics for the three datasets \textit{Negation}, \textit{Generics}, and \textit{ROCStories}, with the initial number of sentences and the explained sentences after filtering. Our filtering strategy removes sentences with more than 15 tokens because of the computational cost, sentences parsed into multiple spans, and sentences containing punctuations \verb|!"#$%&'()*+, -./:;<=>?@[\]^_`{}~| given by the Python module \texttt{string}.

\begin{table}[h]
    \centering
    \small
    \begin{tabular}{lccc}
        \Xhline{3\arrayrulewidth}
         & \textbf{Negation} & \textbf{Generics} & \textbf{ROCStories} \\\hline
        Initial size & 534 & 5777 & 2275 \\
        GPT-2 filter & 534 & 3568 & 1592\\
        Mistral filter & 534 & 3328 & 1543\\
        \Xhline{3\arrayrulewidth}
    \end{tabular}
    \caption{Dataset statistics.}
    \label{tab:dataset_size}
\end{table}

\autoref{fig:n_token_count} displays the length distribution of sentences in each dataset after filtering. \textit{Negation} dataset contains short sentences with less than 8 tokens. It is also used in our study for experiments on coherency and semantic alignment of explanations. \textit{Generics} and \textit{ROCStories} are more complex and realistic. They also have a greater diversity of words and syntactic complexity as identified by the dependency distance and token diversity in \autoref{tab:dataset_descriptives}.

\section{Additional results}

\begin{figure*}
\begin{subfigure}{\textwidth}
    \includegraphics[width=\textwidth]{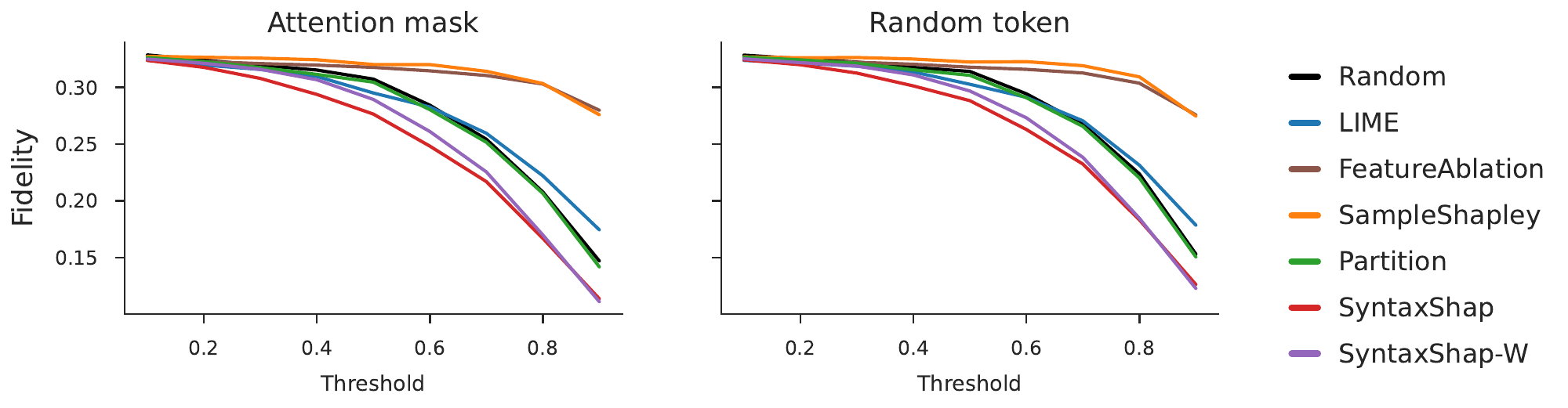}
    \caption{Mistral7B, Generics dataset.}
    \label{fig:mistral_generics}\vspace{1em}
\end{subfigure}
\hfill
\begin{subfigure}{\textwidth}
    \includegraphics[width=\textwidth]{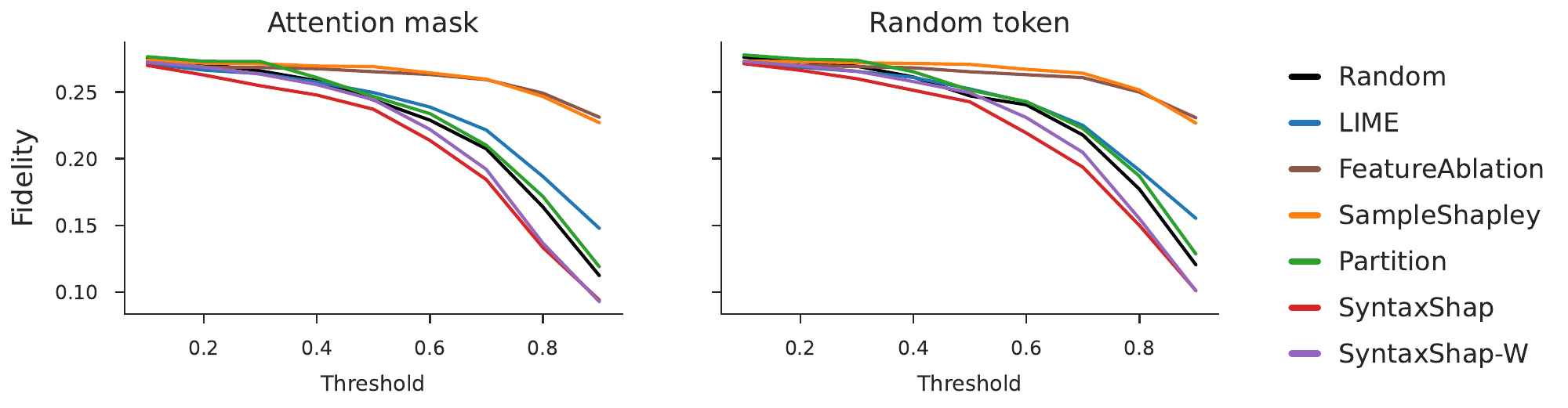}
    \caption{Mistral7B, ROCStories dataset.}
    \label{fig:mistral_ROCStories}
\end{subfigure}
\caption{Masking strategies. The faithfulness of the explanations of Mistral7B predictions by the methods Random, LIME, FeatureAblation, SampleShapley, Partition, and our methods SyntaxShap and SyntaxShap-W, the weighted variant.}
\label{fig:masking_strategies_mistral}
\end{figure*}

\begin{figure*}
\begin{subfigure}{\textwidth}
    \includegraphics[width=\textwidth]{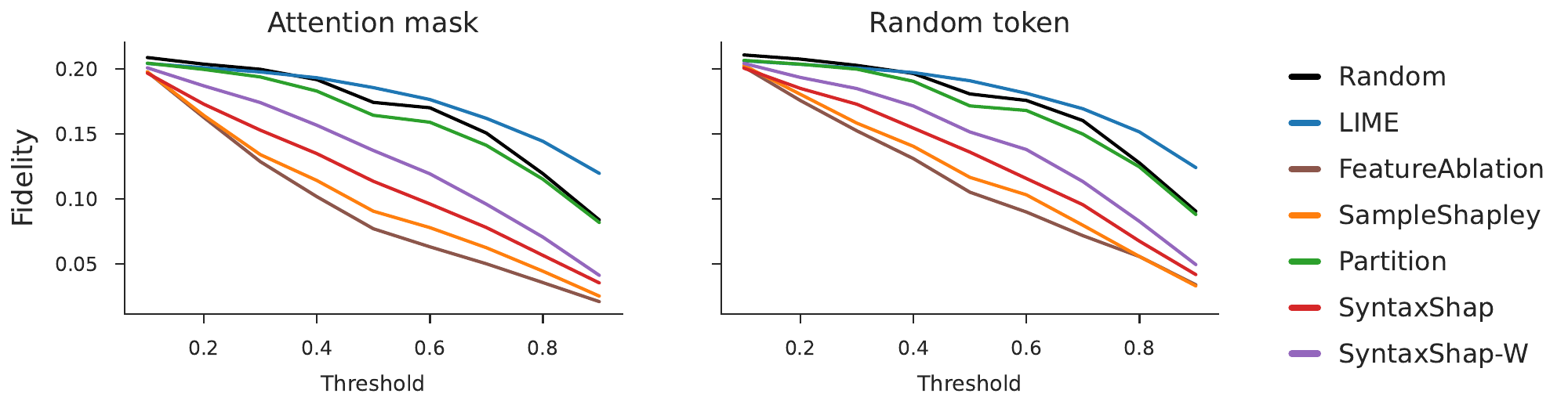}
    \caption{GPT-2, Generics dataset.}
    \label{fig:gpt2_generics}\vspace{1em}
\end{subfigure}
\hfill
\begin{subfigure}{\textwidth}
    \includegraphics[width=\textwidth]{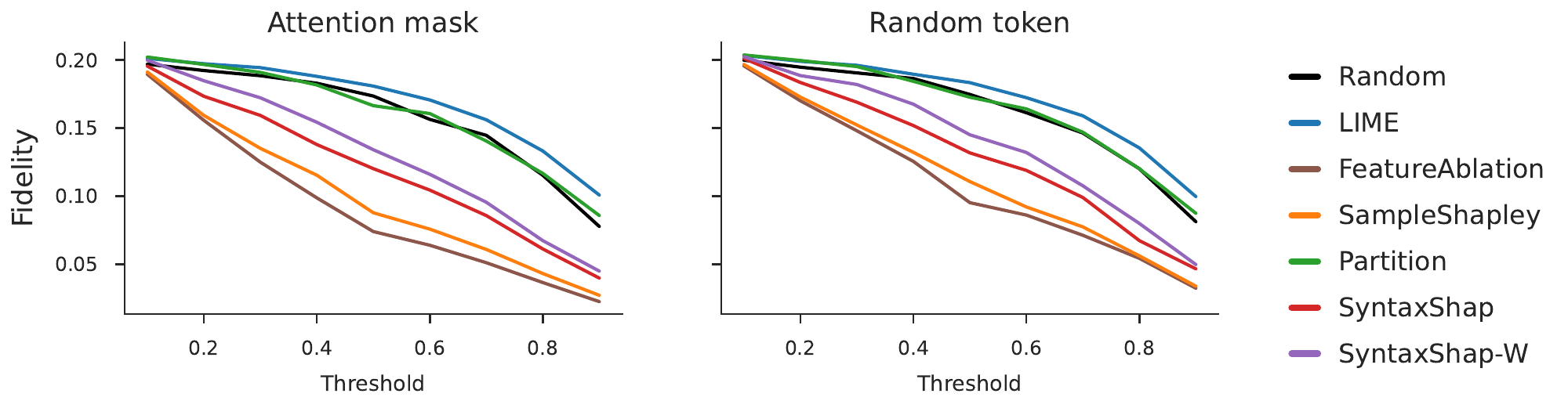}
    \caption{GPT-2, ROCStories dataset.}
    \label{fig:gpt2_ROCStories}
\end{subfigure}
\caption{Masking strategies. The faithfulness of the explanations of GPT-2 predictions by the methods Random, LIME, FeatureAblation, SampleShapley, Partition, and our methods SyntaxShap and SyntaxShap-W, the weighted variant.}
\label{fig:masking_strategies_gpt2}
\end{figure*}

\begin{figure*}[t]
\includegraphics[width=\linewidth]{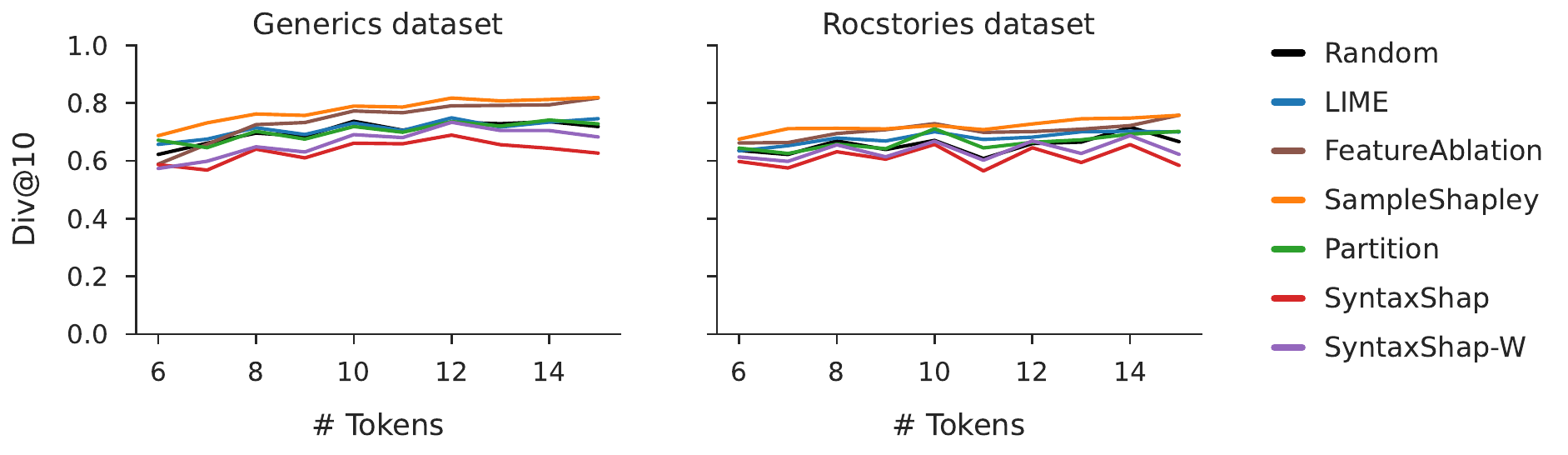}
\caption{Performance of the explainers for Mistral 7B model when varying the number of tokens from 6 to 15.}
\label{fig:n_tokens_mistral}
\end{figure*}

\begin{figure*}[t]
\includegraphics[width=\linewidth]{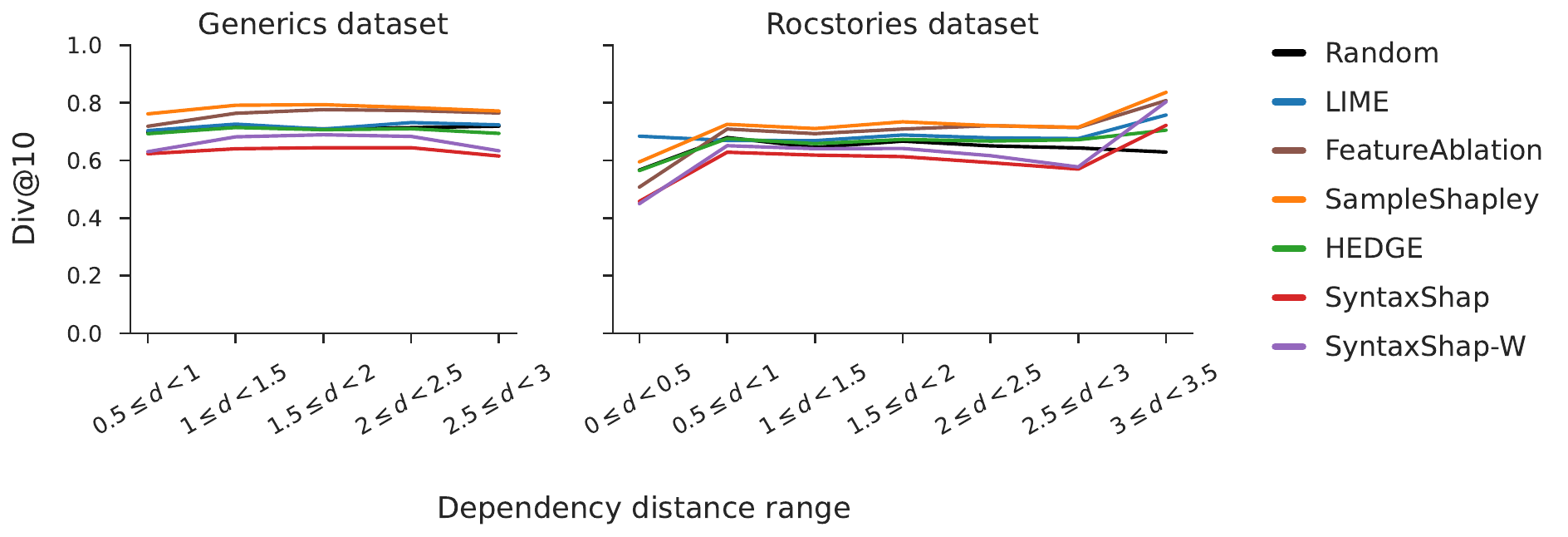}
\caption{Performance of the explainers for Mistral 7B model with respect to the dependency distance varying from $0$ and $3.5$.}
\label{fig:dpdcy_dist_mistral}
\end{figure*}

\subsection{Masking strategies}\label{apx:masking}

This section analyses the impact of selecting different masking strategies when removing unimportant tokens on the faithfulness of explanations. We want to investigate if there is an optimal masking strategy for explainability with text data.
In our evaluation with faithfulness metrics, we experimented with two masking strategies, including (i) modifying attention weights and (ii) replacing tokens with random selections from the tokenizer vocabulary. The left figures in \autoref{fig:masking_strategies_mistral} and \autoref{fig:masking_strategies_gpt2} show the faithfulness scores of explanations when unimportant are masked with null attention, while the right figures display the scores when they are replaced with random tokens. For Mistral 7B model, we observe little difference between the two masking strategies. For GPT-2 model, replacing unimportant tokens with random tokens produces slightly less faithful explanations than removing attention from those tokens. This aligns with our understanding of LLM predictions, as their predictions rely on contextualized embeddings. Random tokens can alter the sentence context, impacting the meaning of the important tokens in the explanation for the model. However, both masking strategies ultimately yield the same conclusions, as the relative performance of explainability methods remains consistent in both cases.

\subsection{Number of tokens and faithfulness}\label{apx:n_tokens}

This section analyzes the relationship between the number of tokens in the input sentences and the performance of the explainability algorithms. We vary the number of tokens from 5 to 15 tokens to have at least 50 sentences of the same length for both \textit{Generics} and \textit{ROCStories} and have a decent number of inputs to average upon (see the number of tokens distribution in \autoref{fig:n_token_count}). \autoref{fig:n_tokens_mistral} shows that the performance of all methods is robust to the increase in the number of tokens. SyntaxShap can be applied to a diverse range of sentence lengths.

\subsection{Dependency distance and faithfulness}

\begin{table}[]
    \centering
    \begin{tabular}{lcc}
    \Xhline{3\arrayrulewidth}
        Depend. dist. & \textbf{Generics}  & \textbf{ROCStories} \\\hline
        $0<d\leq 0.5$ & 0 & 3\\
        $0.5<d\leq 1$ & 624 & 150\\
        $1<d\leq 1.5$ & 1462 & 535\\
        $1.5<d\leq 2$ & 976 & 580\\
        $2<d\leq 2.5$ & 226 & 210\\
        $2.5<d\leq 3$ & 40 & 59\\
        $3<d\leq 3.5$ & 0 & 7\\
        \Xhline{3\arrayrulewidth}
    \end{tabular}
    \caption{Number of input sentences grouped by dependency distance range. Data was filtered with Mistral 7B model.}
    \label{tab:dpdcy_dist}
\end{table}

\begin{figure}
    \centering
    \includegraphics[width=\linewidth]{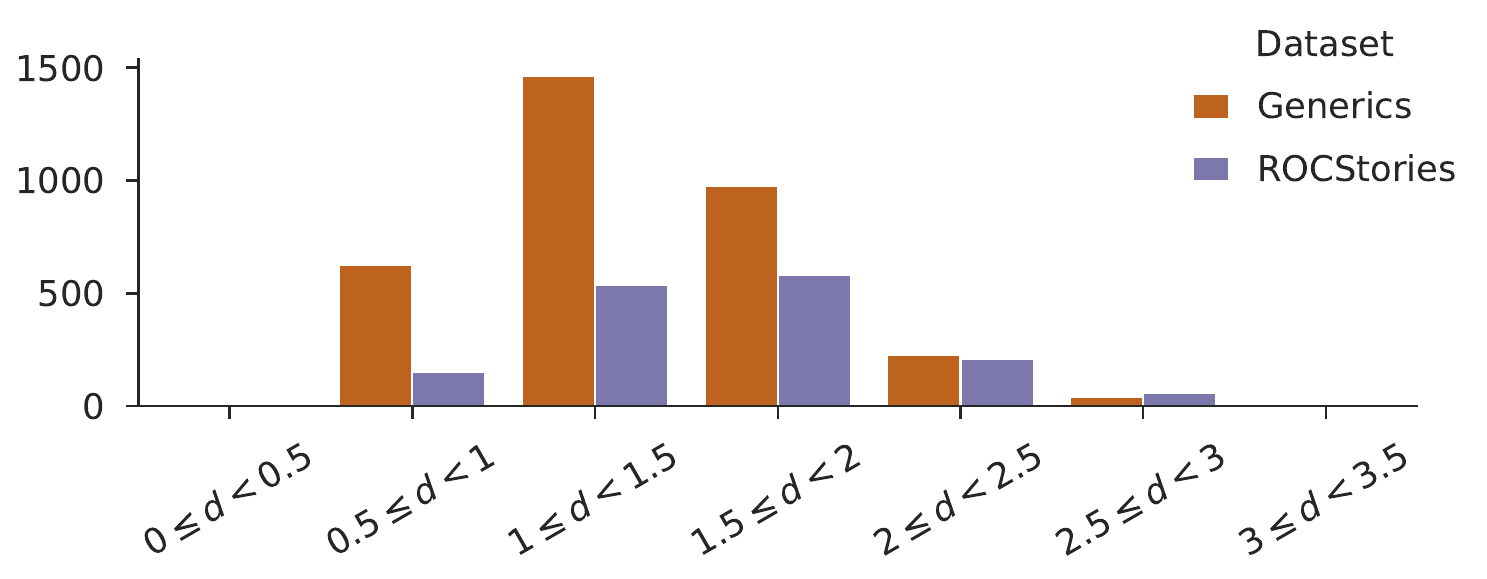}
    \caption{Dependency distance for Generics and ROCStories after filtering with Mistral 7B model.}
    \label{fig:dpdcy_stats_mistral}
\end{figure}

This section explores how the faithfulness of explanations varies with respect to the dependency distance of the input sentence. The average dependency distance (ADD) of a sentence is a good indicator of its syntactic complexity. ADD is calculated by summing the distances of all dependencies in the sentence and then dividing that sum by the number of dependencies in the sentence \cite{dependencydistance}. The distance of a single dependency corresponds to the distance between the two words involved in that dependency within the sentence. We use the TextDescriptives component in \texttt{spaCy} to measure the dependency distance of the analyzed sentences following the universal dependency relations established by~\citet{DependencyTree}. We aim to investigate whether higher syntactic complexity correlates with less faithful explanations, meaning that it may be more challenging for explainability methods to generate faithful explanations. \autoref{fig:dpdcy_dist_mistral} shows that the faithfulness of explanations remains constant across all ranges of dependency distance for the Generics dataset. For ROCStories, we observe slightly less faithful explanations when the input sentences exhibit more complex syntactic structures. However, the number of instances with a low or high dependency distance, as reported in \autoref{tab:dpdcy_dist}, is too low to conclude, with 3 input sentences with a dependency distance $d<0.5$ and 7 with $d>3$. \autoref{fig:dpdcy_stats_mistral} also presents the distribution of sentences grouped by dependency distance range in the ROCStories and Generics datasets. For ROCStories, most input sentences have a dependency distance between 1 and 2. Therefore, observations made for extreme dependency distance ranges in \autoref{fig:dpdcy_dist_mistral} should be interpreted with caution. We do not have enough instances to conclude whether higher dependency distances imply less faithful explanations.

\end{document}